%% file: main.tex
\documentclass[final]{cvpr}

\usepackage{times}
\usepackage{epsfig}
\usepackage{graphicx}
\usepackage{amsmath}
\usepackage{amssymb}
\usepackage{multirow}
\usepackage{booktabs}
\usepackage{color}

\usepackage{marvosym}
\usepackage{lipsum}

\usepackage{appendix}

\newtheorem{definition}{Definition}

\newcommand{\best}[1]{\textbf{\color{blue}#1}}
\newcommand{\comment}[1]{}

\newcommand\blfootnote[1]{%
	\begingroup
	\renewcommand\thefootnote{}\footnote{#1}%
	\addtocounter{footnote}{-1}%
	\endgroup
}

\newcommand{\executeiffilenewer}[3]{%
	\ifnum\pdfstrcmp{\pdffilemoddate{#1}}%
	{\pdffilemoddate{#2}}>0%
	{\immediate\write18{#3}}\fi%
}

\graphicspath{{./}{section/figure/}{images/}}

\usepackage[pagebackref=true,breaklinks=true,colorlinks,bookmarks=false]{hyperref}

\def\cvprPaperID{4113} 
\def\confYear{CVPR 2021}

\begin{document}


\title{OTCE: A Transferability Metric for Cross-Domain Cross-Task Representations}

\author{Yang Tan, Yang Li\textsuperscript{\Letter}, Shao-Lun Huang\\
Tsinghua-Berkeley Shenzhen Institute, Tsinghua University\\
{\tt\small tany19@mails.tsinghua.edu.cn, \{yangli, shaolun.huang\}@sz.tsinghua.edu.cn}
}

\maketitle


\blfootnote{\textsuperscript{\Letter} Corresponding author. This research is funded by Natural Science Foundation of China 62001266.}
\blfootnote{Published in CVPR2021.}

\begin{abstract}
Transfer learning across heterogeneous data distributions (a.k.a.  domains) and distinct tasks is a more general and challenging problem than conventional transfer learning, where either domains or tasks are assumed to be the same. While neural network based feature transfer is widely used in transfer learning applications, finding the optimal transfer strategy still requires time-consuming  experiments and domain knowledge. We propose a transferability metric called Optimal Transport based Conditional Entropy (OTCE), to analytically predict the transfer performance for supervised classification tasks in such cross-domain and cross-task feature transfer settings. Our OTCE score characterizes transferability as a combination of domain difference and task difference, and explicitly evaluates them from data in a unified framework. Specifically, we use optimal transport to estimate domain difference and the optimal coupling between source and target distributions, which is then used to derive the conditional entropy of the target task (task difference). Experiments on the largest cross-domain dataset DomainNet and Office31 demonstrate that OTCE shows an average of $21\%$ gain in the correlation with the ground truth transfer accuracy compared to state-of-the-art methods. We also investigate two applications of the OTCE score including source model selection and multi-source feature fusion.
\end{abstract}

\input{section/introduction}

\input{section/relatedwork}
\input{section/method}
\input{section/experiment}
\input{section/conclusion}

{\small
\bibliographystyle{ieee_fullname}
\bibliography{egbib}
}

\appendix

\section*{Appendix}
\input{appendix/appendix}

\end{document}

%% file: section/introduction.tex
\section{Introduction}
\label{sec: introduction}

\input{section/figure/fig_illustration_transfer_setting}

\textit{Transfer learning} is a useful learning paradigm to improve the performance on target tasks with the help of related source tasks (or source models), especially when only few labeled target data are available for supervision~\cite{pratt1993discriminability,sun2019meta}. A \textit{Transferability} metric can quantitatively reveal how easy it is to transfer knowledge learned from a source task to the target task~\cite{eaton2008modeling,ammar2014automated,sinapov2015learning,LEEP}. It indeed provides a road map for conducting transfer learning in practice, e.g., selecting highly transferable tasks for joint training~\cite{zamir2018taskonomy}, or understanding task relationships for source model selection~\cite{bao2019information,achille2019task2vec,NCE,LEEP}.\par

While theoretical results in transfer learning such as  \cite{ben2010theory,ben2003exploiting,blitzer2008learning,mansour2009domain} suggest that task relationship can be modeled by  certain divergence between the source and target data generating distributions, they are difficult to estimate in practice when target training data is limited. Previous transferability metrics~\cite{zamir2018taskonomy, yosinski2014transferable,achille2019task2vec} empirically calculate the task relationships indicated by training loss or validation accuracy, thus they need to retrain the source model involving expensive computation. Recent analytical metrics~\cite{bao2019information, NCE, LEEP} are limited by strict assumptions on data. NCE~\cite{NCE} assumes that source and target tasks share the same input instances; H-score~\cite{bao2019information} assumes that source and target data are distributed in the same domain. Although LEEP~\cite{DomainNet} does not make any assumptions on source and target data except for having the same input size, it does not work sufficiently well under the cross-domain setting.\par

In this paper, we investigate the transferability estimation problem for classification tasks under the more challenging ~\textit{cross-domain cross-task} setting, as illustrated in Figure \ref{fig:illustration transfer setting}. For most transfer learning problems we encounter in practice, we cannot assume the source and target data are generated from the same distribution (domain) since domain gaps commonly exist in real life due to different acquisition devices and different physical environments. Meanwhile, we also cannot always assume no task difference exists in a transfer learning application as in \textit{transductive domain adaptation}~\cite{pan2009survey}, i.e., source and target tasks have the same category set. We emphasize that the ~\textit{cross-domain cross-task} setting is more challenging compared to previous settings that require shared input data or same domain, since both \textit{domain difference} and \textit{task difference} deteriorate the transfer performance~\cite{pan2009survey, bao2019information,pan2010domain}. \par

To this end, we propose a novel cross-domain cross-task transferability metric called the \textbf{O}ptimal \textbf{T}ransport based \textbf{C}onditional \textbf{E}ntropy, abbreviated as \textbf{OTCE} score. On one hand, compared to the empirical methods~\cite{zamir2018taskonomy, yosinski2014transferable,achille2019task2vec} that need to retrain the source model using gradient descent to estimate the empirical transfer error, our metric is more efficient (about 75x faster) to compute. On the other hand, our OTCE score explicitly learns the \textit{domain difference} and \textit{task difference} in a unified framework, providing a more interpretable result compared to recent analytical metrics~\cite{LEEP,NCE,bao2019information}.\par

More specifically, we measure the  domain difference between source and target data using Wasserstein distance computed by solving the classic Optimal Transport (OT)~\cite{kantorovich1942translocation,peyre2019computational} problem. The OT problem also estimates the joint probability between source and target samples, which allows us to derive the task difference in terms of the conditional entropy between the source and target task labels. Finally, we learn a linear model of transfer accuracy on domain difference  and  task difference, drawing transfer experience  evaluated on a few auxiliary tasks. Albeit its simplicity, the learned model makes it easier to decompose transferability into different factors through model coefficients.\par

Extensive experiments on the largest cross-domain dataset DomainNet~\cite{DomainNet} and Office31~\cite{Office31} demonstrate that our OTCE score shows significantly higher correlation with transfer accuracy, i.e., predicting the transfer performance more accurately with an average of $21\%$ gain compared to state-of-the-art metrics~\cite{LEEP,NCE,bao2019information}. In addition, we further investigate two applications of transferability in source model selection and multi-source feature fusion. In summary, our contributions are follows:\par
 
1) To our knowledge, we are the first to analytically investigate the transferability estimation problem for supervised classification tasks under the more general and challenging cross-domain cross-task setting. \par
2) We propose a novel cross-domain cross-task transferability metric OTCE score which can explicitly evaluate \textit{domain difference} and \textit{task difference} in a unified framework, and predict the transfer performance in advance.\par
3) We show consistent superior performance in predicting transfer performance compared to state-of-the-art metrics and also investigate the applications of OTCE score in source model selection and multi-source feature fusion.  

%% file: section/figure/fig_illustration_transfer_setting.tex
\begin{figure}[t]
    \centering
    \includegraphics[width=0.8\linewidth]{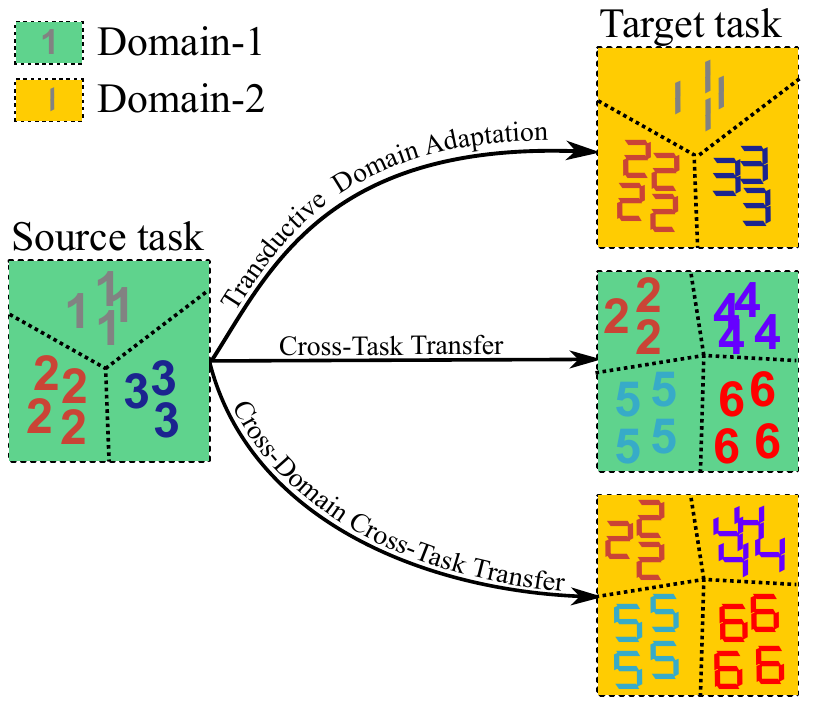}
    \caption{Illustration of three different transfer learning scenarios, i.e., transductive domain adaptation~\cite{pan2009survey}, cross-task transfer~\cite{bao2019information,LEEP} and the cross-domain cross-task transfer we investigating. We take the digital number classification as an example, where the cross-domain cross-task transfer setting suffers both domain difference and task difference.}
    \label{fig:illustration transfer setting}
    \vspace{-0.5cm}
\end{figure}

%% file: section/relatedwork.tex
\section{Related work}
\label{sec:related work}

Our work is closely related to three fields in the transfer learning area~\cite{pan2009survey,lu2020knowledge}, i.e., empirical studies on transferability, analytical studies on transferability and task relatedness.\par

\textbf{Empirical studies on transferability.} Taskonomy~\cite{zamir2018taskonomy} pioneers the investigation in empirically building a taxonomy of tasks. They retrain the source model on each target task and evaluate transfer performance to build up a non-parametric transferablity score called `task affinity'. Task2Vec~\cite{achille2019task2vec} embeds tasks into a low-dimensional vector space so that transferability can be measured using a non-symmetric distance metric. Task2Vec also needs to retrain the large-scale probe network on target task, and then compute the Fisher information matrix to obtain embedding vector. Ying \etal~\cite{ying2018transfer} propose to learn transfer skills from previous transfer learning experiences, and then apply such skills for future target tasks. Generally, these empirical methods involve expensive computation for training, which is mostly avoided in our approach. \par 

\textbf{Analytical studies on transferability.} The advantage of analytical methods is computational efficiency. Previous H-score~\cite{bao2019information} is an information-theoretic approach for analytically evaluating transferability through solving a HGR maximum correlation problem. They focus on the \textit{task transfer learning} problem, which assumes the same input domain among tasks. NCE~\cite{NCE} adopts conditional entropy to evaluate transferability and task hardness under a particular setting, i.e., source and target tasks share the same input instances but different labels. They provide a derivation that the empirical transferability is lower bounded by the negative conditional entropy. The recently proposed LEEP~\cite{LEEP} score is a more general metric compared to the previous two methods. It is defined by the  average log-likelihood of the expected empirical predictor, which predicts the dummy label distributions for target data in source label space and then compute the empirical conditional distribution of target label given the dummy source label. 
In general, these methods either have strict assumptions on data or do not work sufficiently well in a cross-domain setting.\par
\textbf{Task relatedness.} 
Although some theoretic analysis of generalization bounds~\cite{maurer2009transfer,ben2010theory,ben2003exploiting,ben2006analysis,blitzer2008learning,mansour2009domain} in transfer learning and multi-task learning have shed insights on transferability estimation, it is difficult to verify whether their assumptions are satisfied in practical data and even more difficult to compute exactly. Meanwhile, such distance metrics including $\mathcal{F}$-relatedness~\cite{ben2003exploiting}, $\mathcal{A}$-distance~\cite{ben2006analysis} and discrepancy distance~\cite{mansour2009domain} are symmetric while the transferability metric should be non-symmetric since transferring from one task to another is different from transferring in the reverse direction~\cite{LEEP}. Besides, a recent work \cite{alvarez2020geometric} using Optimal Transport (OT) to evaluate dataset distances also measures task relatedness to some extent. Task relatedness is also studied in multi-task learning, since weakly related tasks may worsen the performance compared to single task learning. \cite{jou2016deep,ranjan2017hyperface,rothe2015dex} utilize some prior human knowledge to join highly related tasks. Other works~\cite{kokkinos2017ubernet,misra2016cross} are capable of dynamically adjusting the relatedness of tasks during the training phase. 

%% file: section/method.tex
\section{Transferability Measure via OTCE}
\label{sec: transferability measure via OTCE}
In this section, we start with presenting the transferability definition of classification tasks. Then we introduce some preliminary of Optimal Transport (OT) before  detailing our proposed OTCE score.

\subsection{Transferability Definition}

 Formally, suppose we have a pair of source and target classification tasks whose data are $D_s = \{(x^i_s, y^i_s)\}_{i=1}^m \sim P_s(x,y)$ and $D_t = \{(x^i_t, y^i_t)\}_{i=1}^n \sim P_t(x,y)$ respectively, where $x^i_s, x^i_t \in \mathcal{X}$ and $ y^i_s \in \mathcal{Y}_s, y^i_t \in \mathcal{Y}_t$. Note that $x^i_s, x^i_t \in \mathcal{X}$ only implies  source and target instances have the same input dimension, but they still reside in different domains, i.e. 
 $P_s\neq P_t$. In addition, we are given a source model $(\theta, h_s)$ pre-trained on source data $D_s$, in which $\theta: \mathcal{X}\rightarrow\mathbb{R}^d$ represents a feature extractor producing $d$-dimensional features and $h_s:\mathbb{R}^d \rightarrow \mathcal{P}(\mathcal{Y}_s)$ is the head classifier outputting the final probability distribution of the labels, where $\mathcal{P}(\mathcal{Y}_s)$ is the space of all probability distributions over $\mathcal{Y}_s$.\par
 
Our transferability definition is based on a popular form of neural network based transfer learning (illustrated in Figure \ref{fig: illustration of retrainhead and finetune}), known as \textit{Retrain head}. It keeps the weights of source feature extractor $\theta$ frozen and retrains a new head classifier $h_t$ for target task~\cite{donahue2014decaf,sharif2014cnn,zeiler2014visualizing,oquab2014learning}. The \textit{ground-truth} of transferability can be represented by the empirical transfer performance on the target task, i.e., retrain the source model on target data and then evaluate the classification accuracy on its testing set. We can define the empirical transferability as follows.
\begin{definition}
	The empirical transferability from source task $S$ to target task $T$ is measured by the expected accuracy of the retrained $(\theta, h_t)$ on the testing set of target task:
	\begin{equation}
		\mathrm{Trf}(S \rightarrow T) = \mathbb{E} \left[ \mathrm{acc}(y_t, x_t; \theta, h_t) \right],
		\label{eq:true-transferability}
	\end{equation}
	which indicates how well the source model $\theta$ performs on target task $T$. \cite{NCE} 
\end{definition}\par
 
Although empirical transferability can be the golden standard of describing how easy it is to transfer knowledge from a source task to a target task, it is computationally expensive to obtain. Therefore,   {\em analytical transferability} aims to effectively approximate empirical transferability, without relying on training a new network.\par
 
It is worth mentioning that another type of transfer learning is referred as \textit{Finetune}~\cite{agrawal2014analyzing,girshick2014rich}, i.e., update the feature extractor $\theta$ and the new head classifier $h_t$ simultaneously. Compared to \textit{Retrain head}, \textit{Finetune} trade-offs transfer efficiency for better target accuracy and it requires more target data to avoid overfitting. As in previous analytical studies~\cite{LEEP,NCE,bao2019information}, in this paper, we pay more attention to the \textit{Retrain head} method by working directly in the feature space determined by the source feature extractor $\theta$.  Thus the performance of  current analytical transferability metrics on the finetuned model $(\theta_t, h_t)$ are generally worse than that of {\em Retrain head} for the same tasks. Nevertheless, experiments under the \textit{Finetune} setting (Section \ref{evaluation on few-shot setting}) show that our OTCE score outperforms previous metrics~\cite{LEEP,NCE,bao2019information}.\par   
 
\input{section/figure/fig_retrainhead_finetune}

\subsection{Preliminary of Optimal Transport}
Optimal Transport (OT) theory originated from the Monge problem in 1781, and then the Kantorovich relaxation~\cite{kantorovich1942translocation} was proposed to make the Optimal Transport theory a powerful approach to leverage the underlying space for comparing distributions, shapes and point clouds~\cite{peyre2019computational}. The OT problem considers a complete and separable metric space $\mathcal{X}$, along with continuous or discrete probability measures $\alpha,\beta \in \mathcal{P(X)}$~\cite{alvarez2020geometric}. The Kantorovich relaxation of OT problem is defined as:
\begin{equation}
	OT(\alpha, \beta) \triangleq \mathop{\min}\limits_{\pi \in \Pi(\alpha, \beta)}  \int_{\mathcal{X} \times \mathcal{X}} c(x,z)d\pi (x,z), 
\label{eq: OT definition}
\end{equation}
where $c(\cdot,\cdot): \mathcal{X}\times\mathcal{X} \rightarrow \mathbb{R^+}$ is a cost function, and $\Pi(\alpha,\beta)$ is a set of couplings, i.e., joint probability distributions over the space $\mathcal{X}\times\mathcal{X}$ with marginal distributions $\alpha, \beta$, that is,
\begin{equation}
	\Pi(\alpha,\beta) \triangleq \{ \pi \in \mathcal{P(X \times X)}| P_{1\#}\pi=\alpha, P_{2\#}\pi=\beta \}.
\end{equation} 
When the $c(x,z)=d_\mathcal{X}(x,z)^p$ of some $p\geq1$, $W_p(\alpha,\beta) \triangleq OT(\alpha,\beta)^{1 / p }$ is denoted as the p-Wasserstein distance.\par

In practice, we  rarely  know the true marginal distributions  $\alpha,\beta$. Instead, we usually compute the discrete empirical distributions $\hat{\alpha}=\sum_{i=1}^{m}\mathbf{a}_i\delta_{\mathbf{x}^i}, \hat{\beta}=\sum_{i=1}^{n}\mathbf{b}_i\delta_{\mathbf{z}^i}$, where $\mathbf{a},\mathbf{b}$ are vectors in the probability simplex. And the cost function in Equation \eqref{eq: OT definition} can simply be represented as an $m\times n$ cost matrix $\mathbf{C}$, where $\mathbf{C}_{ij}=c(\mathbf{x}^i, \mathbf{z}^j)$.\par

Furthermore, OT can be efficiently solved via the Sinkhorn algorithm~\cite{cuturi2013sinkhorn} by adding an entropic regularizer to the objective function in Equation \eqref{eq: OT definition}.  The entropic regularized OT has been used for domain adaptation to compute the optimal mapping of input from the source domain to the target domain \cite{courty2016optimal}. Alvarez \etal~\cite{alvarez2020geometric} also adopts OT to geometrically evaluate the distance between datasets.

\subsection{OTCE Score}

The motivation of our OTCE (Optimal Transport based Conditional Entropy) score is decomposing the overall difference between two classification tasks into \textit{domain difference} and \textit{task difference}. 
 To this end, we adopt OT to evaluate the domain difference for its advantages in computing directly from finite empirical samples and capturing the underlying geometry of data. More importantly, by solving the OT problem between source and target data, we can obtain an optimal coupling matrix of samples, revealing the pair-wise optimal matching under a given distance metric.\par 

\input{section/figure/fig_markov}

From a probabilistic point of view, the coupling matrix is a non-parametric estimation of the joint probability of the source and target latent features $P(X_s,X_t)$. We model the relationship between the source and the target data according to the following simple Markov random field: $Y_s-X_s-X_t-Y_t$ (shown in Figure \ref{fig: markov}), where label random variables $Y_s$ and $Y_t$ are only dependent on  $X_s$ and $X_t$, respectively, i.e, $P(Y_s,Y_t|X_s,X_t) = P(Y_s|X_s)P(Y_t|X_t)$.
Furthermore, we can derive the empirical joint probability distribution of source and target label sets,
\begin{equation}
P(Y_s,Y_t) = \mathbb{E}_{X_s,X_t}[P(Y_s|X_s)P(Y_t|X_t)].
\label{eq: joint label distribution}
\end{equation}
 We consider this joint probability distribution to some extent represents the task difference, since the goodness of class-to-class matching may intuitively reveal the hardness of transfer. Inspired by Tran \etal~\cite{NCE} who use Conditional Entropy (CE) $H(Y_t|Y_s)$ to describe class-to-class matching quality over the same input instances, we consider it as a reasonable metric to evaluate task difference once we learn the soft correspondence between source and target features $P(X_s,X_t)$ via optimal transport. Finally, we define our analytical transferability metric OTCE as a weighted combination of domain difference and task difference.\par

\input{section/figure/fig_otce_pipeline}

Figure \ref{fig: illustration of OTCE} shows the overview of our proposed transferability metric. The computation process of OTCE score is described in following steps.\par
 \textbf{Step1: Compute domain difference.} In our problem, we adopt the OT definition with entropic regularization~\cite{cuturi2013sinkhorn} to facilitate the computation:
\begin{equation}
\begin{aligned}
	OT(D_s, D_t)  \triangleq  \mathop{\min}\limits_{\pi \in \Pi(D_s, D_t)}  \sum_{i,j=1}^{m,n}  c(\theta(x^i_s),\theta(x^j_t))\pi_{ij} + \epsilon H(\pi), 
\end{aligned}
\label{eq: OT definition with entropy}
\end{equation}
where $c(\cdot, \cdot) = \| \cdot - \cdot \| ^2_2$ is the cost metric, and $\pi$ is the coupling matrix of size $m\times n$, and $H(\pi)=-\sum_{i=1}^m \sum_{j=1}^n \pi_{ij}\log\pi_{ij}$ is the entropic regularizer with $\epsilon=0.1$. The OT problem above can be solved efficiently by Sinkhorn algorithm~\cite{cuturi2013sinkhorn} to produce an optimal coupling matrix $\pi^*$. Thus the \textit{domain difference} $W_D$ can be represented by the commonly used 1-Wasserstein distance, denoted as:
\begin{equation}
W_D =  \sum_{i,j=1}^{m,n}  \| \theta(x^i_s)-\theta(x^j_t)\|^2_2\pi_{ij}^*.
	\label{eq: domain difference}
\end{equation}\par

 \textbf{Step2: Compute task difference.} Based on the optimal coupling matrix $\pi^*$, we can compute the empirical joint probability distribution of source and target label sets, and the marginal probability distribution of source label set, denoted as: 
 \begin{equation}
 	\hat{P}(y_s,y_t) = \sum_{i,j: y^i_s=y_s, y^j_t=y_t} \pi_{ij}^*,
 	\label{eq: joint distribution}
 \end{equation}

 \begin{equation}
	\hat{P}(y_s) = \sum_{y_t \in \mathcal{Y}_t} \hat{P}(y_s,y_t).
	\label{eq: marginal distribution}
\end{equation}
Note that Equation \eqref{eq: joint distribution} is the empirical estimation of Equation \eqref{eq: joint label distribution} for all pairs of source and target samples.    
Then we can compute the Conditional Entropy (CE) to represent \textit{task difference} $W_T$, 
 \begin{equation}
\begin{aligned}
	W_T & = H(Y_t|Y_s) = H(Y_s,Y_t) - H(Y_s)\\
	           & = - \sum_{y_t \in \mathcal{Y}_t} \sum_{y_s \in \mathcal{Y}_s} \hat{P}(y_s,y_t)\log \frac{\hat{P}(y_s,y_t)}{\hat{P}(y_s)}.
\end{aligned}
	\label{eq: task difference}
\end{equation}\par

Here we explain why CE can be used to measure task difference. As the testing accuracy of the target task can be well indicated by the training log-likelihood score $l_T(\theta,h_t)$ if the model is not overfitted, we can define an alternative  empirical transferability metric to Equation \eqref{eq:true-transferability} as follows:
 \begin{equation}
	\begin{aligned}
\widetilde{\mathrm{Trf}}(S \rightarrow T) & = l_T(\theta, h_t) \\
& =  \frac{1}{n}\sum^n_{i=1}\log P(y^i_t|x^i_t; \theta,h_t).
	\end{aligned}
	\label{eq: alternative definition of true transferability}
\end{equation}
So the following relationship is obtained,
\begin{equation}
		\widetilde{\mathrm{Trf}}(S \rightarrow T) \ge l_S(\theta, h_s) - H(Y_t|Y_s).
	\label{eq: transferability inequality}
\end{equation}
Proof is detailed in \cite{NCE}. $l_S(\theta, h_s)$ is a constant after the source model training, so the lower bound of transferability is determined by the Conditional Entropy (CE). In other words, the empirical transferability can be attributed to CE. \par
However, in cross-domain setting, CE alone is not sufficient to estimate empirical transferability as discussed in the Appendix Section \ref{appendix: study of ot-based nce}. One reason could be that there exists inherent uncertainty in estimating the joint distribution of source and target features through empirical samples, so we need to capture such uncertainty through domain difference. Thus the following step is to combine \textit{domain difference} and \textit{task difference} to obtain our OTCE score.\par

 \textbf{Step3: Compute OTCE score.} Intuitively, we model the OTCE score as a linear combination of \textit{domain difference} and \textit{task difference}:
 \begin{equation}
 	OTCE=\lambda _1 \hat{W}_D + \lambda_ 2\hat{W}_T + b,
 	\label{eq: OTCE score}
 \end{equation}
 where $\lambda_1, \lambda_2$ are weighting coefficients for standardized \textit{domain difference} $\hat{W}_D$ and \textit{task difference} $\hat{W}_T$ respectively, and $b$ is the bias term. Choosing the optimal weights is a challenging task since the importance of $\hat{W}_D$ and $\hat{W}_T$ may be different for various cross-domain configurations, as described in Section \ref{study of auxiliary task number}.\par

 Consequently, we learn the coefficients for current specified source and target domains utilizing several auxiliary tasks. More specifically, we sample several pairs of source and target tasks, and compute their \textit{domain differences, task differences} and \textit{empirical transferability} as the transfer experience. Least square fitting is used to obtain the adjusted $\lambda_1, \lambda_2, b$. While the OTCE score can be generalized to higher order polynomial, we favor linear model since it is fast to compute (with analytical solution) and more interpretable. After obtaining the fitted model, we can use the OTCE score to predict transfer accuracy for any source-target task pair in the current cross-domain setting. 

%% file: section/figure/fig_retrainhead_finetune.tex
\begin{figure}[t]
\centering
\def\svgwidth{0.6\linewidth}
	\executeiffilenewer{images/retrainhead_finetune.svg}{images/retrainhead_finetune.pdf}%
	{inkscape -z -D --file=images/retrainhead_finetune.svg %
		--export-pdf=images/retrainhead_finetune.pdf --export-latex}%
	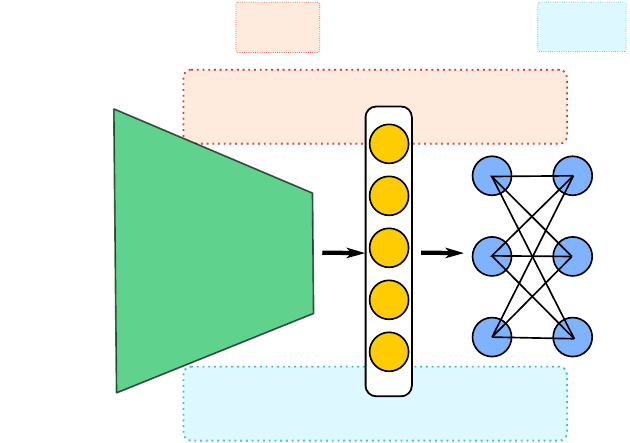%

\caption{Illustration of two neural network based transfer learning methods, i.e., Retrain head and Finetune.}
\label{fig: illustration of retrainhead and finetune}
\vspace{-0.4cm}
\end{figure}

%% file: images/retrainhead_finetune.pdf_tex
\begingroup%
  \makeatletter%
  \providecommand\color[2][]{%
    \errmessage{(Inkscape) Color is used for the text in Inkscape, but the package 'color.sty' is not loaded}%
    \renewcommand\color[2][]{}%
  }%
  \providecommand\transparent[1]{%
    \errmessage{(Inkscape) Transparency is used (non-zero) for the text in Inkscape, but the package 'transparent.sty' is not loaded}%
    \renewcommand\transparent[1]{}%
  }%
  \providecommand\rotatebox[2]{#2}%
  \newcommand*\fsize{\dimexpr\f@size pt\relax}%
  \newcommand*\lineheight[1]{\fontsize{\fsize}{#1\fsize}\selectfont}%
  \ifx\svgwidth\undefined%
    \setlength{\unitlength}{181.41732554bp}%
    \ifx\svgscale\undefined%
      \relax%
    \else%
      \setlength{\unitlength}{\unitlength * \real{\svgscale}}%
    \fi%
  \else%
    \setlength{\unitlength}{\svgwidth}%
  \fi%
  \global\let\svgwidth\undefined%
  \global\let\svgscale\undefined%
  \makeatother%
  \begin{picture}(1,0.70312499)%
    \lineheight{1}%
    \setlength\tabcolsep{0pt}%
    \put(0,0){\includegraphics[width=\unitlength,page=1]{retrainhead_finetune.pdf}}%
    \put(0.31109064,0.27887666){\makebox(0,0)[lt]{\lineheight{1.64999998}\smash{\begin{tabular}[t]{l}\large{$\theta$}\end{tabular}}}}%
    \put(0.56,0.55172831){\makebox(0,0)[lt]{\lineheight{1.64999998}\smash{\begin{tabular}[t]{l}$\theta(x_t)$\end{tabular}}}}%
    \put(0.96193193,0.27167141){\makebox(0,0)[lt]{\lineheight{1.64999998}\smash{\begin{tabular}[t]{l}$h_t$\end{tabular}}}}%
    \put(0.07301812,0.31886237){\makebox(0,0)[lt]{\lineheight{1.64999998}\smash{\begin{tabular}[t]{l}$x_t$\\\end{tabular}}}}%
    \put(0,0){\includegraphics[width=\unitlength,page=2]{retrainhead_finetune.pdf}}%
    \put(0.00,0.63797235){\makebox(0,0)[lt]{\lineheight{1.64999998}\smash{\begin{tabular}[t]{l}Retrain head:\end{tabular}}}}%
    \put(0,0){\includegraphics[width=\unitlength,page=3]{retrainhead_finetune.pdf}}%
    \put(0.58,0.63797235){\makebox(0,0)[lt]{\lineheight{1.64999998}\smash{\begin{tabular}[t]{l}Finetune:\end{tabular}}}}%
    \put(0,0){\includegraphics[width=\unitlength,page=4]{retrainhead_finetune.pdf}}%
    \put(0.05883239,0.015){\makebox(0,0)[lt]{\lineheight{1.64999998}\smash{\begin{tabular}[t]{l}\small{Frozen}\end{tabular}}}}%
    \put(0,0){\includegraphics[width=\unitlength,page=5]{retrainhead_finetune.pdf}}%
    \put(0.0588324,0.08){\makebox(0,0)[lt]{\lineheight{1.64999998}\smash{\begin{tabular}[t]{l}\small{Free}\end{tabular}}}}%
  \end{picture}%
\endgroup%

%% file: section/figure/fig_markov.tex
\begin{figure}
\centering
\vspace{-1.3cm}
\def\svgwidth{0.7\linewidth}
	\executeiffilenewer{images/markov.svg}{images/markov.pdf}%
	{inkscape -z -D --file=images/markov.svg %
		--export-pdf=images/markov.pdf --export-latex}%
	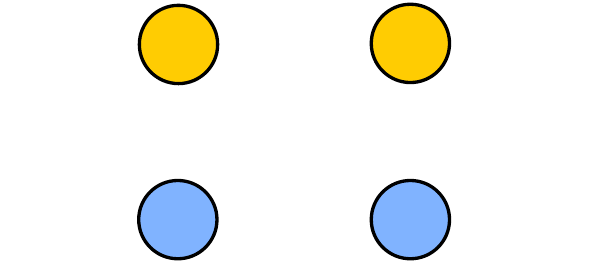
\caption{The probability graph model of the source and target task. 
}
\label{fig: markov}
\vspace{-0.4cm}
\end{figure}

%% file: images/markov.pdf_tex
\begingroup%
  \makeatletter%
  \providecommand\color[2][]{%
    \errmessage{(Inkscape) Color is used for the text in Inkscape, but the package 'color.sty' is not loaded}%
    \renewcommand\color[2][]{}%
  }%
  \providecommand\transparent[1]{%
    \errmessage{(Inkscape) Transparency is used (non-zero) for the text in Inkscape, but the package 'transparent.sty' is not loaded}%
    \renewcommand\transparent[1]{}%
  }%
  \providecommand\rotatebox[2]{#2}%
  \newcommand*\fsize{\dimexpr\f@size pt\relax}%
  \newcommand*\lineheight[1]{\fontsize{\fsize}{#1\fsize}\selectfont}%
  \ifx\svgwidth\undefined%
    \setlength{\unitlength}{141.73228346bp}%
    \ifx\svgscale\undefined%
      \relax%
    \else%
      \setlength{\unitlength}{\unitlength * \real{\svgscale}}%
    \fi%
  \else%
    \setlength{\unitlength}{\svgwidth}%
  \fi%
  \global\let\svgwidth\undefined%
  \global\let\svgscale\undefined%
  \makeatother%
  \begin{picture}(1,0.65999999)%
    \lineheight{1}%
    \setlength\tabcolsep{0pt}%
    \put(0,0){\includegraphics[width=\unitlength,page=1]{markov.pdf}}%
    \put(0.27,0.35){\makebox(0,0)[lt]{\lineheight{1.64999998}\smash{\begin{tabular}[t]{l}$Y_s$\end{tabular}}}}%
    \put(0.26,0.05){\makebox(0,0)[lt]{\lineheight{1.64999998}\smash{\begin{tabular}[t]{l}$X_s$\end{tabular}}}}%
    \put(0.66,0.05){\makebox(0,0)[lt]{\lineheight{1.64999998}\smash{\begin{tabular}[t]{l}$X_t$\end{tabular}}}}%
    \put(0.67,0.35){\makebox(0,0)[lt]{\lineheight{1.64999998}\smash{\begin{tabular}[t]{l}$Y_t$\end{tabular}}}}%
    \put(0,0){\includegraphics[width=\unitlength,page=2]{markov.pdf}}%
    \put(0.02,0.2){\makebox(0,0)[lt]{\lineheight{1.64999998}\smash{\begin{tabular}[t]{l}$P(X_s,Y_s)$\end{tabular}}}}%
    \put(0.72,0.2){\makebox(0,0)[lt]{\lineheight{1.64999998}\smash{\begin{tabular}[t]{l}$P(X_t,Y_t)$\end{tabular}}}}%
    \put(0.36,0.12){\makebox(0,0)[lt]{\lineheight{1.64999998}\smash{\begin{tabular}[t]{l}$P(X_s,X_t)$\end{tabular}}}}%
  \end{picture}%
\endgroup%

%% file: section/figure/fig_otce_pipeline.tex
\begin{figure}[t]
\centering
\def\svgwidth{1.0\linewidth}
	\executeiffilenewer{images/OTCE.svg}{images/OTCE.pdf}%
	{inkscape -z -D --file=images/OTCE.svg %
		--export-pdf=images/OTCE.pdf --export-latex}%
	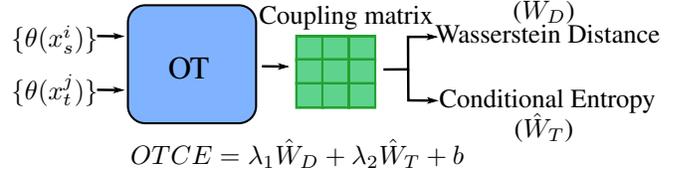%

\caption{Illustration of our proposed OTCE transferability metric.}
\label{fig: illustration of OTCE}
\vspace{-0.4cm}
\end{figure}

%% file: images/OTCE.pdf_tex
\begingroup%
  \makeatletter%
  \providecommand\color[2][]{%
    \errmessage{(Inkscape) Color is used for the text in Inkscape, but the package 'color.sty' is not loaded}%
    \renewcommand\color[2][]{}%
  }%
  \providecommand\transparent[1]{%
    \errmessage{(Inkscape) Transparency is used (non-zero) for the text in Inkscape, but the package 'transparent.sty' is not loaded}%
    \renewcommand\transparent[1]{}%
  }%
  \providecommand\rotatebox[2]{#2}%
  \newcommand*\fsize{\dimexpr\f@size pt\relax}%
  \newcommand*\lineheight[1]{\fontsize{\fsize}{#1\fsize}\selectfont}%
  \ifx\svgwidth\undefined%
    \setlength{\unitlength}{233.85826772bp}%
    \ifx\svgscale\undefined%
      \relax%
    \else%
      \setlength{\unitlength}{\unitlength * \real{\svgscale}}%
    \fi%
  \else%
    \setlength{\unitlength}{\svgwidth}%
  \fi%
  \global\let\svgwidth\undefined%
  \global\let\svgscale\undefined%
  \makeatother%
  \begin{picture}(1,0.26666667)%
    \lineheight{1}%
    \setlength\tabcolsep{0pt}%
    \put(0,0){\includegraphics[width=\unitlength,page=1]{OTCE.pdf}}%
    \put(0.25,0.14502897){\makebox(0,0)[lt]{\lineheight{1.64999998}\smash{\begin{tabular}[t]{l}\large{OT}\end{tabular}}}}%
    \put(0.000,0.195){\makebox(0,0)[lt]{\lineheight{1.64999998}\smash{\begin{tabular}[t]{l}$\{\theta(x^i_s)\}$\end{tabular}}}}%
    \put(0.00,0.11182469){\makebox(0,0)[lt]{\lineheight{1.64999998}\smash{\begin{tabular}[t]{l}$\{\theta(x^j_t)\}$\end{tabular}}}}%
    \put(0,0){\includegraphics[width=\unitlength,page=2]{OTCE.pdf}}%
    \put(0.395,0.23153893){\makebox(0,0)[lt]{\lineheight{1.64999998}\smash{\begin{tabular}[t]{l}Coupling matrix\end{tabular}}}}%
    \put(0,0){\includegraphics[width=\unitlength,page=3]{OTCE.pdf}}%
    \put(0.67604287,0.20105761){\makebox(0,0)[lt]{\lineheight{1.64999998}\smash{\begin{tabular}[t]{l}Wasserstein Distance\end{tabular}}}}%
    \put(0.68222286,0.09909875){\makebox(0,0)[lt]{\lineheight{1.64999998}\smash{\begin{tabular}[t]{l}Conditional Entropy\end{tabular}}}}%
    \put(0.79989837,0.24){\makebox(0,0)[lt]{\lineheight{1.64999998}\smash{\begin{tabular}[t]{l}($\hat{W}_D$)\end{tabular}}}}%
    \put(0.80202378,0.05){\makebox(0,0)[lt]{\lineheight{1.64999998}\smash{\begin{tabular}[t]{l}($\hat{W}_T$)\end{tabular}}}}%
    \put(0.19,0.008344){\makebox(0,0)[lt]{\lineheight{1.64999998}\smash{\begin{tabular}[t]{l}$OTCE = \lambda_1 \hat{W}_D  + \lambda_2  \hat{W}_T + b$\end{tabular}}}}%
  \end{picture}%
\endgroup%

%% file: section/experiment.tex
\section{Experiments}
\label{sec: experiment}

We first evaluate our OTCE score on the largest-to-date cross-domain dataset, DomainNet~\cite{DomainNet}, and  another popular dataset Office31~\cite{Office31} by computing the   Pearson correlation coefficient between OTCE score and the empirical transferability (transfer accuracy). Three different transfer settings are considered, namely the \textit{standard setting},
the \textit{fixed category set size setting}, and the \textit{few-shot setting}. Then we investigate the applications of our transferability metric in source model selection and multi-source feature fusion. Finally, we study the effects of the number of auxiliary task on the performance of our transferability metric. More discussions are included in the Appendix.

\subsection{Datasets}
\label{subsec: datasets} 
We generate collections of classification tasks by sampling different sets of categories from two existing cross-domain image datasets:\par

\textbf{DomainNet~\cite{DomainNet}} contains six domains (styles) of images, i.e., \textit{Clipart (C), Infograph (I), Painting (P), Quickdraw (Q), Real (R) and Sketch (S)}, each covering 345 common object categories. We exclude the \textit{Infograph} domain due to its noisy annotations. It is worth mentioning that categories in DomainNet are severely imbalanced, i.e., the number of images per category ranges from 8 to 586. To eliminate the influence of imbalanced data in obtaining the empirical transferability, we limit the number of instances per category to be at most $\leq 100$ in all target tasks.\par

\textbf{Office31~\cite{Office31}} is a common benchmark dataset for domain adaptation algorithms on three domains: \textit{Amazon (A), DSLR (D)} and \textit{ Webcam (W)}. It contains 4,110 images of 31 categories of objects typically found in an office environment.

\subsection{Evaluation on Standard Setting}
\label{evaluation on standard setting}

\input{section/table/tab_transferability_vs_acc}

\input{section/figure/fig_correlation}

We define a standard cross-domain cross-task setting to evaluate the correlation (measured by Pearson correlation coefficient like~\cite{LEEP,NCE}) between our proposed OTCE score and the transfer accuracy. We compare performances with recent analytical transferability metrics LEEP~\cite{LEEP}, NCE~\cite{NCE} and H-score~\cite{bao2019information}. As the original NCE assumes that the source and target tasks are different labels on  the same instances, we follow the modified implementation by \cite{LEEP}, i.e., use the source model to predict the dummy source label for target data. \par

To generate source tasks, we obtain a 44-category and a 15-category classification tasks for DomainNet and Office31 respectively through random sampling. Then we train 8 source models (5 for DomainNet, 3 for Office31) for different domains on the defined source tasks initialized using an ImageNet-pretrained~\cite{russakovsky2015imagenet} ResNet-18~\cite{he2016deep} model. \par

For target tasks, we randomly sample 100 classification tasks from each target domain. The number of categories  range from 10-100  for DomainNet, and  10-31 for Office31, respectively. In each transfer configuration, we select one domain as the source domain, and consider others as target domains.  Thus in this setting, we totally conduct $5\times4\times100=2000$  cross-domain cross-task transfer tests on DomainNet, and $3\times2\times100=600$ tests on Office31. To determine the coefficients $\lambda_1, \lambda_2, b$ of OTCE, we randomly select $10\%$ target tasks as the auxiliary for each cross-domain configuration (specified by the set of source and target domains involved), and others are used for testing. The empirical transferability of each target task is the testing accuracy after training the source model on target data with SGD optimizer and cross entropy loss for 100 epochs. Table \ref{tab: transferability vs acc} (upper part) shows the numerical results of comparing our proposed OTCE score with LEEP, NCE and H-score. Figure \ref{fig: correlation between acc and transferability} (the first row) visualizes the correlations of transferability metrics and empirical transferability (ground truth) on the test data. Both Table \ref{tab: transferability vs acc} and Figure \ref{fig: correlation between acc and transferability} clearly demonstrate that OTCE score achieves higher correlation with the ground truth across all domain configurations, with about $5\%$, $9\%$ and $27\%$ gain compared to LEEP, NCE and H-score respectively.

\subsection{Evaluation on Fixed Category Set Size}
\label{evaluation on fixed label set size}
Analyzing the experimental results of Section \ref{evaluation on standard setting}, we find that transfer accuracy drops with the increasing of category set size (number of categories), shown in the Appendix Section \ref{appendix: study of category set size}. A larger category set generally makes it more difficult to learn the target task well under the same training setting for a given source model. Such differences in the intrinsic  complexity of the target task  tends to overshadow the more subtle variations in transferability due to task and domain relatedness. To show OTCE score indeed captures these subtle variations, we design a more challenging experiment where all target category set sizes are the same. Specifically, we sample 100 target tasks with $category\_set\_size = 50$ for each target domain, and follow the training strategy described in the standard setting (Section \ref{evaluation on standard setting}). Results shown in Table \ref{tab: transferability vs acc} (middle part) and Figure \ref{fig: correlation between acc and transferability} (the third row) demonstrate that our proposed OTCE score outperforms other transferability metrics by a large margin, with an  average $34\%$ and $39\%$ correlation gain compared to LEEP and NCE respectively. We also note that H-score failed in this difficult setting, since correlation coefficients are negative where they should be positive.

\subsection{Evaluation on Few-shot Setting}
\label{evaluation on few-shot setting}

The few-shot learning problem~\cite{sun2019meta} is a common application scenario of transfer learning, since training from scratch using only few-shot samples (e.g., 10 samples per category) can easily overfit, while transferring representations from a highly related source model can greatly improve the generalization of the target task. Thus it is necessary to test our transferability metric under the few-shot setting. Specifically, few-shot setting is different from the standard setting (Section \ref{evaluation on standard setting}) in two aspects. On one hand, we limit each category only containing 10 samples. On the other hand, we also study the correlation of transferability metrics and the transfer accuracy obtained through \textit{Finetune}. It is worth mentioning that all the aforementioned analytical transferability metrics rely on the feature representation inferred by the source feature extractor. Therefore, it is more challenging to require transferability metrics are still highly correlated with the finetuned accuracy. Despite these restrictions, results shown in Table \ref{tab: transferability vs acc} (lower part) and Figure \ref{fig: correlation between acc and transferability} (the second row) demonstrate that OTCE score is consistently better than LEEP, NCE and H-score with $22\%$, $39\%$ and $83\%$ correlation gain respectively.

 \subsection{Application for Source Model Selection}
\label{application for source model selection}
Selecting the best pretrained source model for a target task from a given set of source models is one of the most common applications of transferability metrics. In this experiment, we adopt 100 target tasks for a specified target domain as in Section \ref{evaluation on standard setting}. And for each target task, there are four source models pretrained on other domains. We want to evaluate whether the source model showing highest transferability score has the highest transfer accuracy on target task. If so, we consider that transferability score successfully predicts the best source model. Finally, we calculate the ratio of successful predictions. We compare the prediction accuracy among OTCE, LEEP and NCE. H-score is omitted since it does not produce any meaningful result in this experiment. Quantitative comparisons shown in Table \ref{tab: model selection} show that our OTCE score achieves top results in predicting the best source model. 

\input{section/table/tab_model_selection}

\subsection{Application for Multi-Source Feature Fusion} 
\label{application for feature fusion}

We test OTCE score on a multi-source feature fusion problem, which is another application scenario  of transferability, i.e., one can transfer multiple source models to a target task by merging their inferred features together to obtain a fused representation~\cite{hou2017dualnet}. A simple but effective fusion approach is  element-wise addition or concatenation of source features. A new head classifier can be trained by taking the fused representation as input to produce the final output. However, different source models may result in different transfer performance on the target task. Thus simple average fusion is unable to effectively exploit the most useful information provided by source models. Consequently, we apply the OTCE score to weight the feature fusion for better transfer performance.\par
   
In this experiment, we sample 50 target tasks in \textit{Real} domain of DomainNet dataset from the few-shot setting (Section \ref{evaluation on few-shot setting}). Then we employ 4 source models trained on other domains respectively to perform feature fusion targeting to these target tasks. We use a softmax function to normalize the OTCE scores of four source models to obtain the fusion coefficients in range $[0,1]$, and then multiply source features respectively. We consider two methods of merging features, i.e., element-wise addition and concatenation. Results shown in Figure \ref{fig:feature fusion} demonstrate that feature fusion weighted by our OTCE score achieves the highest testing accuracy on target tasks as expected. Heuristically, our proposed transferability metric OTCE score can be an effective tool for multi-source transfer learning.

\input{section/figure/fig_feature_fusion}

\input{section/figure/fig_study_auxiliary_task}

\subsection{Number of Auxiliary Tasks}
\label{study of auxiliary task number}

Auxiliary tasks are used to determine the coefficients $\lambda_1, \lambda_2$ and $b$ in Equation \eqref{eq: OTCE score} through least square fitting. We analyze the effect of auxiliary tasks on OTCE correlation using DomainNet. As shown in Figure \ref{fig:study of auxiliary task ratio}, we plot the correlation between OTCE score and empirical transferability against the number of auxiliary tasks among all target tasks in each transfer setting. Note that the first data point $number = 0$ (i.e. no auxiliary training) represents the correlation using the pre-defined coefficients $\lambda_1=\lambda_2=-0.5$. This experiment demonstrates that learning the coefficients with auxiliary tasks for different cross-domain setting is necessary to maintain the robustness of OTCE score. Nevertheless, we can still achieve high correlation performance using only few auxiliary tasks. Moreover, we further discuss only using domain difference or task difference to characterize transferability and analyze the learned coefficients in the Appendix Section \ref{appendix: analysis on learned coefficients}. 

%% file: section/table/tab_transferability_vs_acc.tex
\begin{table*}[!t]
\begin{minipage}{\linewidth}

\setlength\tabcolsep{4pt} 

\footnotesize

\caption{Quantitative comparisons evaluated by Pearson correlation coefficients between transferability metrics and transfer accuracy under cross-domain cross-task transfer settings, including standard setting (Section \ref{evaluation on standard setting}), fixed category set size setting (Section \ref{evaluation on fixed label set size}) and few-shot setting (Section \ref{evaluation on few-shot setting}). Superscript $^*$ denotes $p>0.001$. }
\label{tab: transferability vs acc}

\centering

\begin{tabular}{ccccccccc}
\toprule

\multirow{3}{*}{Transferring type}  & \multirow{3}{*}{Dataset}  & \multicolumn{3}{c}{Experimental setting} & \multirow{3}{*}{OTCE} &\multirow{3}{*}{LEEP\cite{LEEP}} & \multirow{3}{*}{NCE\cite{NCE}} & \multirow{3}{*}{H-score\cite{bao2019information}} \\
\cmidrule{3-5} 
& & Source domain & Target domain & Data property &  &  & & \\
\midrule

\multirow{9} {*}{Retrain head} & \multirow{5} {*}{DomainNet} & C & P,Q,R,S & standard & \textbf{0.969} & 0.919 & 0.787 & 0.864  \\
& & P & C,Q,R,S&  standard & \textbf{0.968}  & 0.886 & 0.812 & 0.858 \\
& & Q & C,P,R,S& standard & \textbf{0.963} & 0.942 & 0.935 & 0.843 \\
& & R& C,P,Q,S& standard & \textbf{0.972} & 0.892 & 0.851 & 0.870  \\
& & S & C,P,Q,R& standard & \textbf{0.960} & 0.952 & 0.954& 0.882  \\
\cmidrule{2-9} 
& \multirow{3} {*}{Office31} & A & D,W& standard & \textbf{0.829} & 0.805 & 0.796 & 0.590 \\
& & D & A,W& standard  & \textbf{0.880} & 0.857 & 0.849 & 0.441 \\
& & W & A,D& standard & \textbf{0.863} & 0.811 & 0.804 & 0.489 \\
\cmidrule{5-9} 
& & & & \textbf{average}  & \textbf{0.926} & 0.883 & 0.849 & 0.730  \\

\midrule
\midrule

\multirow{6} {*}{Retrain head} & \multirow{6} {*}{DomainNet} & C & P,Q,R,S & fixed category set size & \textbf{0.757} & 0.614 & 0.535 & -0.599  \\
& & P & C,Q,R,S&  fixed category set size & \textbf{0.712} & 0.480 & 0.418 & -0.541 \\
& & Q & C,P,R,S& fixed category set size & \textbf{0.352} & 0.213 & 0.269 & -0.288 \\
& & R& C,P,Q,S&fixed category set size & \textbf{0.639} & 0.465 & 0.440 & $-0.100^*$  \\
& & S & C,P,Q,R&fixed category set size  & \textbf{0.435} & 0.381 & 0.427& -0.302  \\

\cmidrule{5-9} 
& & & & \textbf{average}  & \textbf{0.579} & 0.431 & 0.418  &  -0.346 \\

\midrule
\midrule

\multirow{8} {*}{Retrain head} & \multirow{5} {*}{DomainNet} & C & P,Q,R,S & few-shot & \textbf{0.920} & 0.843 & 0.713 & 0.767  \\
& & P & C,Q,R,S&  few-shot & \textbf{0.924} & 0.812 & 0.737 & 0.807 \\
& & Q & C,P,R,S& few-shot & \textbf{0.852} & 0.836 & 0.825 & 0.786 \\
& & R& C,P,Q,S& few-shot & \textbf{0.937} & 0.787  & 0.744& 0.814  \\
& & S & C,P,Q,R& few-shot  & \textbf{0.922} & 0.886 & 0.884 & 0.834  \\
\cmidrule{2-9} 
& \multirow{3} {*}{Office31} & A & D,W& few-shot  & \textbf{0.840} & 0.803 & 0.793 & 0.640 \\
& & D & A,W& few-shot   & \textbf{0.933} & 0.923 & 0.930 & 0.413 \\
& & W & A,D& few-shot  & \textbf{0.927} & 0.920 & 0.926 & $0.277^*$ \\

\midrule

\multirow{9} {*}{Finetune} & \multirow{5} {*}{DomainNet} & C & P,Q,R,S & few-shot & \textbf{0.699} & 0.333 & $0.153^*$ & 0.406  \\
& & P & C,Q,R,S&  few-shot & \textbf{0.766} & 0.414 & 0.309 & 0.554 \\
& & Q & C,P,R,S& few-shot & \textbf{0.663} & 0.623 & 0.635 & 0.607 \\
& & R& C,P,Q,S& few-shot & \textbf{0.854} & 0.288 & 0.226 & 0.511  \\
& & S & C,P,Q,R& few-shot  & \textbf{0.681} & 0.514 & 0.526& 0.481  \\
\cmidrule{2-9} 
& \multirow{3} {*}{Office31} & A & D,W& few-shot  & \textbf{0.319} & $0.210^*$ & $0.204^*$ & $0.173^*$ \\
& & D & A,W& few-shot   & \textbf{0.939} & 0.865 & 0.896 & $0.186^*$ \\
& & W & A,D& few-shot  & \textbf{0.947} & 0.875 & 0.883 & $-0.002^*$ \\

\cmidrule{5-9} 
& & & & \textbf{average}  & \textbf{0.820} & 0.670 & 0.627 & 0.476  \\

\bottomrule

\end{tabular}

\end{minipage}
\vspace{-0.3cm}
\end{table*}

%% file: section/figure/fig_correlation.tex
\begin{figure*}[t]
\centering
\def\svgwidth{0.9\textwidth}
	\executeiffilenewer{images/correlation.svg}{images/correlation.pdf}%
	{inkscape -z -D --file=images/correlation.svg %
		--export-pdf=images/correlation.pdf --export-latex}%
	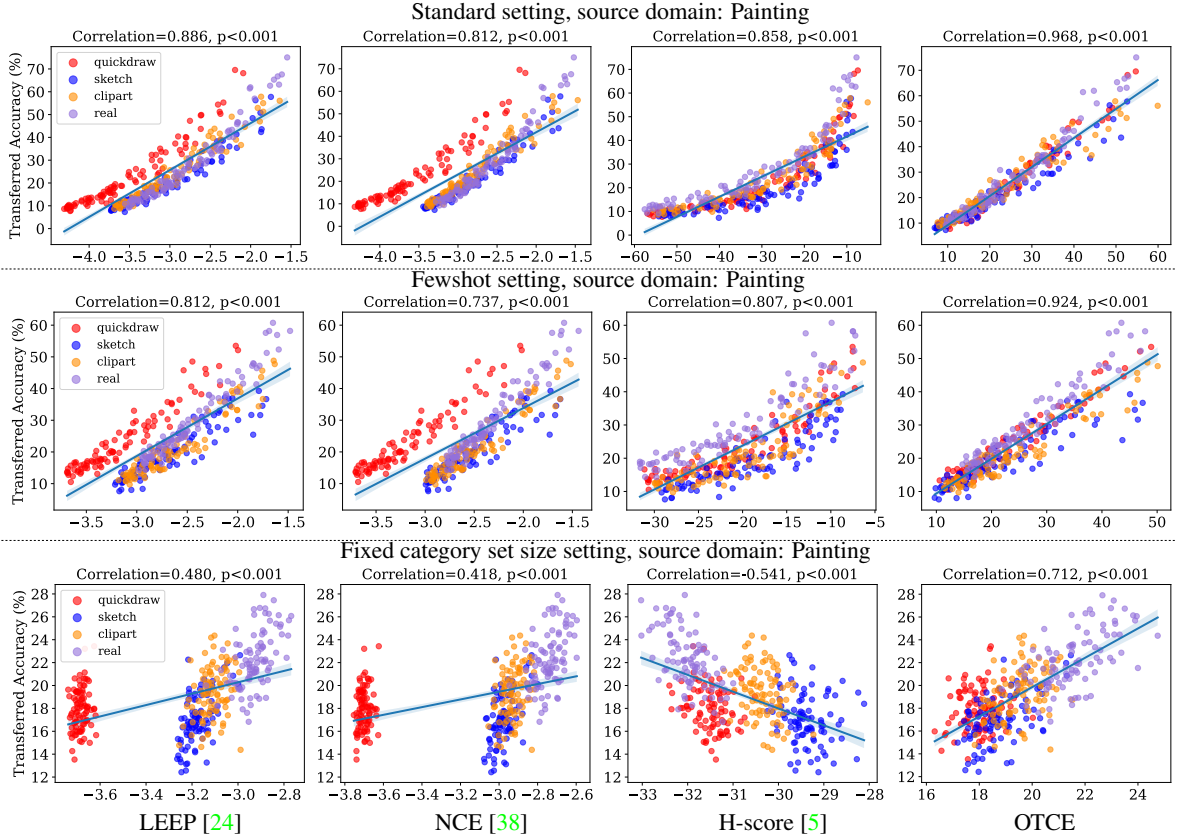%

\caption{Visualization of correlations between empirical transferability (transfer accuracy through Retrain head) and analytical transferability metrics, including LEEP, NCE, H-score and our proposed OTCE score. Each row shows the correlations under standard setting, few-shot setting and fixed category set size setting respectively, where source domain is \textit{Painting} and target domains are \textit{Clipart, Quickdraw, Real, Sketch}. Points in the figure represent different target tasks. It can be seen that our OTCE score shows significantly better correlations with empirical transferability.}
\label{fig: correlation between acc and transferability}
\vspace{-0.4cm}
\end{figure*}

%% file: images/correlation.pdf_tex
\begingroup%
  \makeatletter%
  \providecommand\color[2][]{%
    \errmessage{(Inkscape) Color is used for the text in Inkscape, but the package 'color.sty' is not loaded}%
    \renewcommand\color[2][]{}%
  }%
  \providecommand\transparent[1]{%
    \errmessage{(Inkscape) Transparency is used (non-zero) for the text in Inkscape, but the package 'transparent.sty' is not loaded}%
    \renewcommand\transparent[1]{}%
  }%
  \providecommand\rotatebox[2]{#2}%
  \newcommand*\fsize{\dimexpr\f@size pt\relax}%
  \newcommand*\lineheight[1]{\fontsize{\fsize}{#1\fsize}\selectfont}%
  \ifx\svgwidth\undefined%
    \setlength{\unitlength}{496.06299213bp}%
    \ifx\svgscale\undefined%
      \relax%
    \else%
      \setlength{\unitlength}{\unitlength * \real{\svgscale}}%
    \fi%
  \else%
    \setlength{\unitlength}{\svgwidth}%
  \fi%
  \global\let\svgwidth\undefined%
  \global\let\svgscale\undefined%
  \makeatother%
  \begin{picture}(1,0.70857141)%
    \lineheight{1}%
    \setlength\tabcolsep{0pt}%
    \put(0,0){\includegraphics[width=\unitlength,page=1]{correlation.pdf}}%
    \put(0.12,0.00784208){\makebox(0,0)[lt]{\lineheight{1.64999998}\smash{\begin{tabular}[t]{l}\small{LEEP~\cite{LEEP}}\end{tabular}}}}%
    \put(0.37,0.00784208){\makebox(0,0)[lt]{\lineheight{1.64999998}\smash{\begin{tabular}[t]{l}\small{NCE~\cite{NCE}}\end{tabular}}}}%
    \put(0.61,0.00784208){\makebox(0,0)[lt]{\lineheight{1.64999998}\smash{\begin{tabular}[t]{l}\small{H-score~\cite{bao2019information}}\end{tabular}}}}%
    \put(0.86,0.00784208){\makebox(0,0)[lt]{\lineheight{1.64999998}\smash{\begin{tabular}[t]{l}\small{OTCE}\end{tabular}}}}%
    \put(0,0){\includegraphics[width=\unitlength,page=2]{correlation.pdf}}%
    \put(0.29,0.23754078){\makebox(0,0)[lt]{\lineheight{1.64999998}\smash{\begin{tabular}[t]{l}\small{Fixed category set size setting, source domain: Painting}\end{tabular}}}}%
    \put(0.35,0.69413635){\makebox(0,0)[lt]{\lineheight{1.64999998}\smash{\begin{tabular}[t]{l}\small{Standard setting, source domain: Painting}\end{tabular}}}}%
    \put(0,0){\includegraphics[width=\unitlength,page=3]{correlation.pdf}}%
    \put(0.35,0.46735053){\makebox(0,0)[lt]{\lineheight{1.64999998}\smash{\begin{tabular}[t]{l}\small{Fewshot setting, source domain: Painting}\end{tabular}}}}%
  \end{picture}%
\endgroup%

%% file: section/table/tab_model_selection.tex
\begin{table}[!t]
\begin{minipage}{\linewidth}

\footnotesize

\caption{Quantitative comparisons of source model selection accuracy ($\%$) among transferability metrics on DomainNet.  }
\label{tab: model selection}

\centering

\begin{tabular}{l rrr rrr}
\toprule
\multirow{3} {*}{Method} & \multicolumn{5}{c}{Target domain} & \multirow{3} {*}{average} \\
 \cmidrule{2-6}
 & \multicolumn{1}{c}{C}  & \multicolumn{1}{c}{P} & \multicolumn{1}{c}{Q} & \multicolumn{1}{c}{R} & \multicolumn{1}{c}{S} & \\
\midrule

LEEP\cite{LEEP} & 31.1 & 26.7 & 5.6 & 97.8 & \textbf{100.0} & 52.2\\
NCE\cite{NCE} &\textbf{41.1} & \textbf{94.4} & 2.2 & \textbf{100.0} & \textbf{100.0} & 67.5\\
OTCE & \textbf{41.1} & 93.3 & \textbf{97.8} & \textbf{100.0} & \textbf{100.0} & \textbf{86.4}\\

\bottomrule
 
\end{tabular}

\end{minipage}
\end{table}

%% file: section/figure/fig_feature_fusion.tex
\begin{figure}[t]
    \centering
    \includegraphics[width=0.9\linewidth]{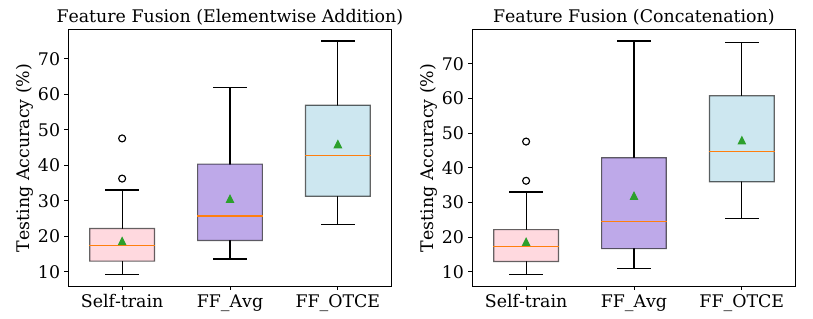}
    \caption{Testing accuracy comparisons among `Self-train' (directly training on target data), `FF\_Avg' (average fusion) and `FF\_OTCE' (fusion weighted by OTCE score).}
    \label{fig:feature fusion}
    \vspace{-0.3cm}
\end{figure}

%% file: section/figure/fig_study_auxiliary_task.tex
\begin{figure}[t]
    \centering
    \includegraphics[width=0.6\linewidth]{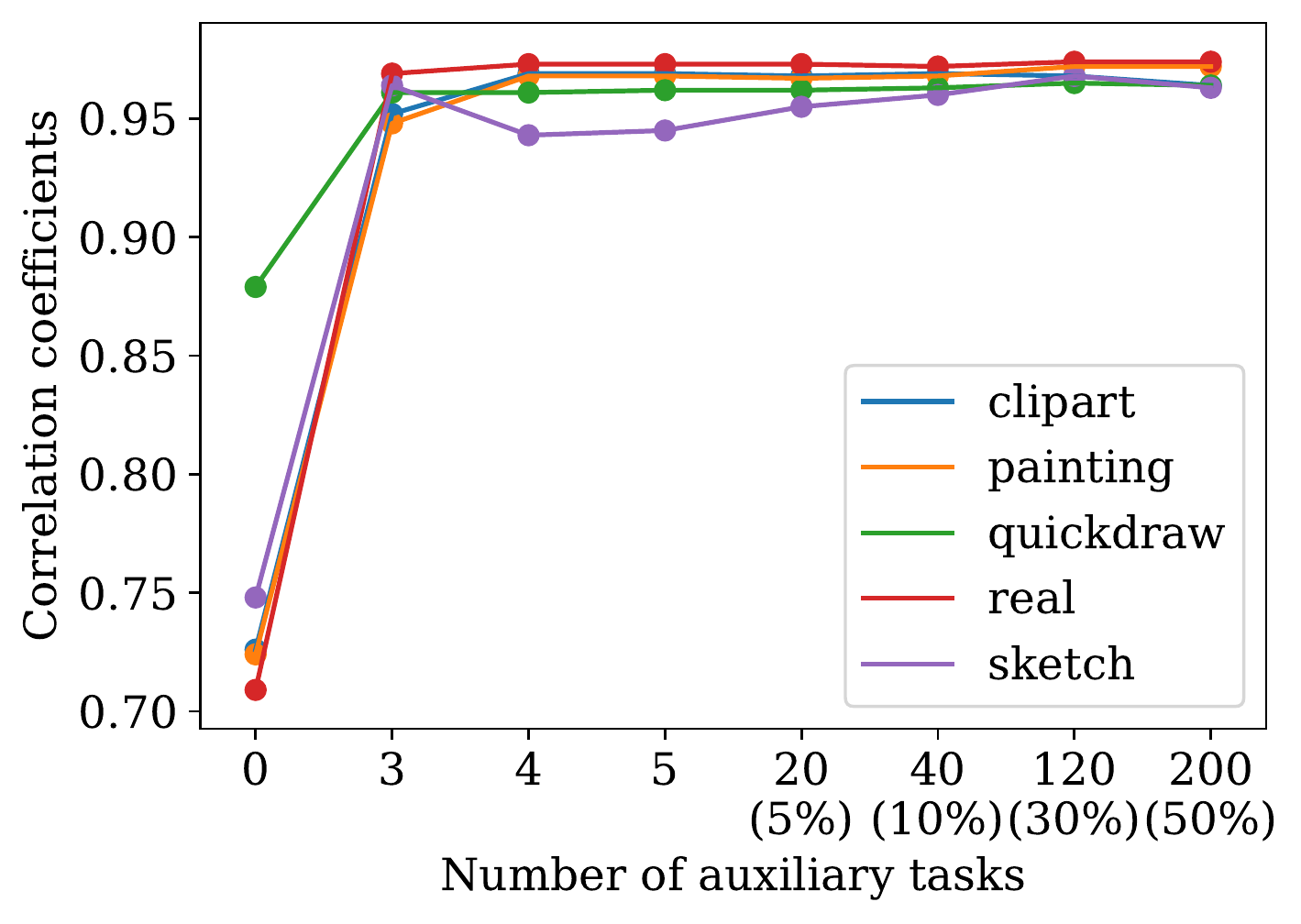}
    \caption{Study of how different number of auxiliary tasks affects the final correlations between OTCE score and empirical transferability. Polygonal lines represent different source domains from DomainNet as described in Section \ref{evaluation on standard setting}. }
    \label{fig:study of auxiliary task ratio}
    \vspace{-0.4cm}
\end{figure}

%% file: section/conclusion.tex
\section{Conclusion}
\label{sec: conclusion}
In this study, we investigated the analytical transferability estimation problem under the general setting of cross-domain cross-task transfer learning. Our proposed transferability metric, OTCE score, characterizes the transferability between source and target tasks based on their \textit{domain difference} and \textit{task difference}, which can be explicitly evaluated in a unified framework. Extensive experiments demonstrate that OTCE score is more robust than other existing analytical transferability methods for capturing the uncertainty in the actual transfer performance under the cross-domain cross-task setting. For applications, we also showed through simple case studies that the OTCE score is a suitable metric to select the best source model in transfer learning and to determine feature weights in multi-source feature fusion for multi-task learning. In future works, we will explore more applications of OTCE score, such as utilizing the domain difference and task difference to support the training procedure in cross-domain cross-task transfer learning problems, e.g. open-set domain adaptation.

%% file: appendix/appendix.tex
\section{Visualization of Datasets }
\label{appendix: visualization of datasets}

\begin{figure}[h]
	\centering
	\includegraphics[width=0.8\linewidth]{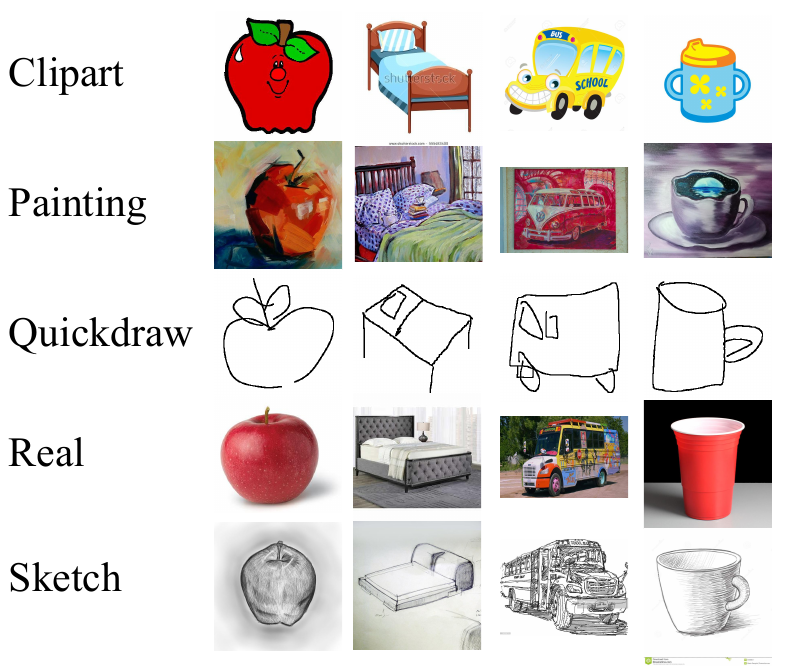}
	\caption{Visualization of data instances from DomainNet~\cite{DomainNet}.}
	\label{fig:visualization of domainnet}
	\vspace{-0.4cm}
\end{figure}

\begin{figure}[h]
	\centering
	\includegraphics[width=0.8\linewidth]{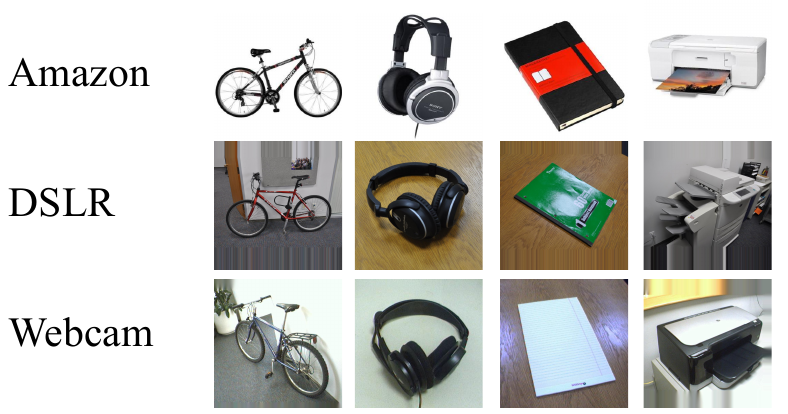}
	\caption{Visualization of data instances from Office31~\cite{Office31}.}
	\label{fig:visualization of office31}
	\vspace{-0.4cm}
\end{figure}

Some data instances of DomainNet~\cite{DomainNet} and Office31~\cite{Office31} are depicted in Figure \ref{fig:visualization of domainnet}, \ref{fig:visualization of office31} respectively. We can see that the five domains of DomainNet have unique image styles and the three domains of Office31 differ in their image acquisition devices and environments. We can also observe that the learning difficulty of images in each domain differs widely. For instance, \textit{Quickdraw} has the least visual complexity among DomainNet's five domains for just having black-and-white lines; Product photos in \textit{Amazon} have far less uncertainty than those in other datasets that contain real world photos for having clean background and simple lighting condition. Such difference is reflected in our transferability analysis using OTCE.

\section{Analysis of OTCE}
\label{appendix: analysis of OTCE}

\begin{figure*}[!t]
    \vspace{-0.5cm}
	\centering
	\includegraphics[width=0.75\textwidth]{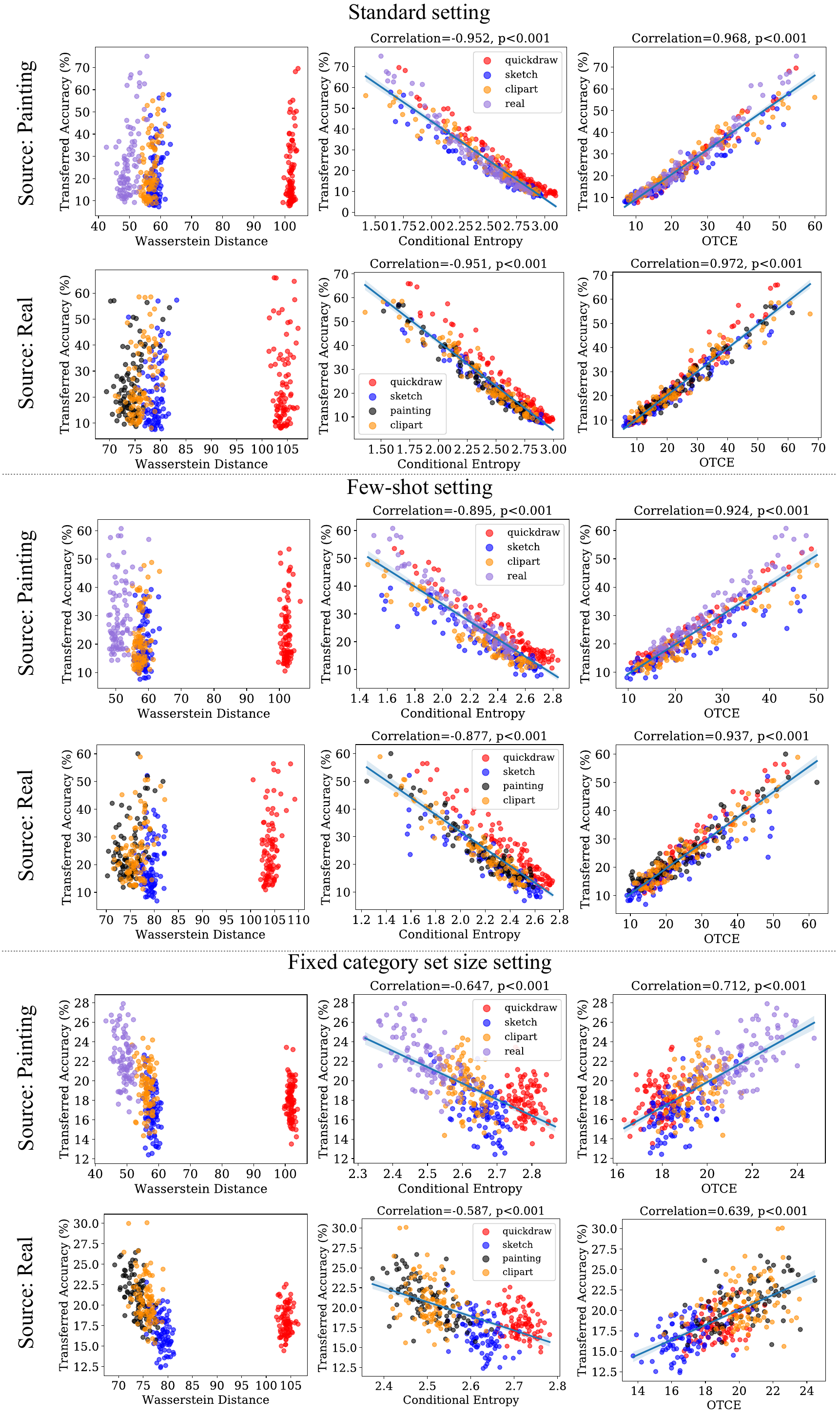}
	\caption{Visualization of OTCE score and its components, i.e., Wasserstein distance (domain difference $W_D$) and Conditional Entropy (task difference $W_T$). Points represent different target tasks.}
	\label{fig:analysis of OTCE}
\end{figure*}
Recall that our OTCE score characterizes transferability based on two factors: \textit{domain difference} $W_D$ measured by cross-domain Wasserstein distance and \textit{task difference} $W_T$ measured by the conditional entropy between the source and target labels under the optimal coupling between two domains. In this section, we examine the relationship between each factor and transferability under the three transfer settings discussed in the paper.\par

\input{appendix/tab_comparison}

First, observing the Wasserstein distances of all transfer instances (left column in Figure \ref{fig:analysis of OTCE}), we find that the domain difference is relatively stable among different target tasks when source and target domains are fixed. This is reasonable since domain difference should be task agnostic. Meanwhile, domain difference shows a generally negative correlation with the transfer accuracy, which is most obvious when the category set size is fixed (row 5-6 of Figure \ref{fig:analysis of OTCE}). The only exception in this case is when the target domain is \textit{Quickdraw} (represented by red points). Due to the low visual complexity in line drawings, most features can achieve relatively high classification accuracy on \textit{Quickdraw} despite being trained on very different source domains. Note that we do not require that the correlation of domain difference and transfer accuracy to be strictly negative. Because in our unified framework, domain difference and task difference are coupled due to the coupling matrix computed via OT.\par

Second, as we emphasize in the paper (Section 3.3) that task difference $W_T$ measured by conditional entropy  alone is not sufficient to characterize cross-domain transferability, we present the experimental evidence to support this finding by looking at the correlation between $W_T$ and the transfer accuracy. In Figure \ref{fig:analysis of OTCE}, our OTCE score (right column) shows significantly higher correlation with the transfer accuracy compared to conditional entropy (middle column). The improvement is most notable under the few-shot and fixed category set size setting, which shows that incorporating domain difference $W_D$ can indeed improve the robustness of transferability prediction in weakly supervised and challenging scenarios.

\section{Study of OT-based NCE}
\label{appendix: study of ot-based nce}

Here we define an alternative transferability metric, \textbf{OT-based NCE} to be the negative conditional entropy ($-W_{T}$) mentioned earlier. Although it is not as robust as OTCE in estimating transferability, it does not require auxiliary task for parameter fitting, and thus is more efficient. We make quantitative comparisons of OT-based NCE with previous transferability metrics LEEP~\cite{LEEP}, NCE~\cite{NCE} and H-score~\cite{bao2019information}. Results shown in Table \ref{tab: OT-based NCE} demonstrate that our OT-based NCE also outperforms previous metrics on average. To conclude, our proposed two transferability metrics, i.e., OTCE and OT-based NCE, possess different advantages and readers can choose flexibly according to their need.

\begin{itemize}
	\item \textbf{OTCE.} It suits the scenario which needs the most accurate transferability estimation or there are many target tasks. In addition, the learned coefficients of domain difference and task difference may benefit some downstream transfer learning applications.
	
	\item \textbf{OT-based NCE.} It is a simple implementation of OTCE, providing relatively coarse but more efficient transferability estimation without extra computation in auxiliary tasks. Although it is not as accurate as OTCE, it still averagely outperforms SOTA analytical metrics.   
	
\end{itemize}

\section{Analysis on Learned Coefficients}
\label{appendix: analysis on learned coefficients}

\begin{figure}[h]
	\centering
	\includegraphics[width=1.0\linewidth]{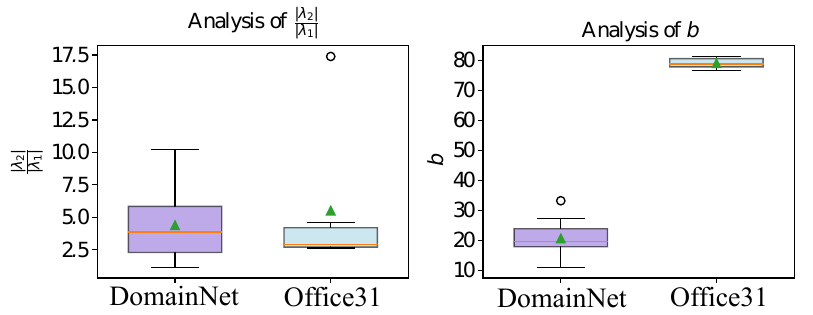}
	\caption{Analysis of learned coefficients. }
	\label{fig: coefficients analysis}
\end{figure}

Setting $\lambda_1 = 1,\lambda_2=0$ or $\lambda_1 = 0,\lambda_2=1$, i.e., only using domain difference or task difference (OT-based NCE) to characterize transferability, does not perform as good as OTCE, as depicted in the Figure \ref{fig:analysis of OTCE} and Table \ref{tab: OT-based NCE}. Moreover, we further analyze the learned coefficients in all experimental settings. We found that bias $b$ was stable for the same cross-domain dataset (shown in Figure \ref{fig: coefficients analysis}), but $\lambda_1, \lambda_2$ among different transfer configurations varied irregularly. On one hand, the importance of domain difference and task difference varies for different cross-domain configurations. On the other hand, differences are calculated in the feature space of source model, so the learned coefficients have different scales among source models. Generally, task difference is more important in describing transferability ($\frac{|\lambda_2|}{|\lambda_1|} > 1$). To conclude, we recommend using auxiliary tasks to learn the coefficients for obtaining the most accurate transferability estimation for the given cross-domain transfer configuration. 

\section{Study of Category Set Size}
\label{appendix: study of category set size}

\begin{figure}[t]
  	\centering
  	\vspace{-17.5cm}
  	\includegraphics[width=0.6\linewidth]{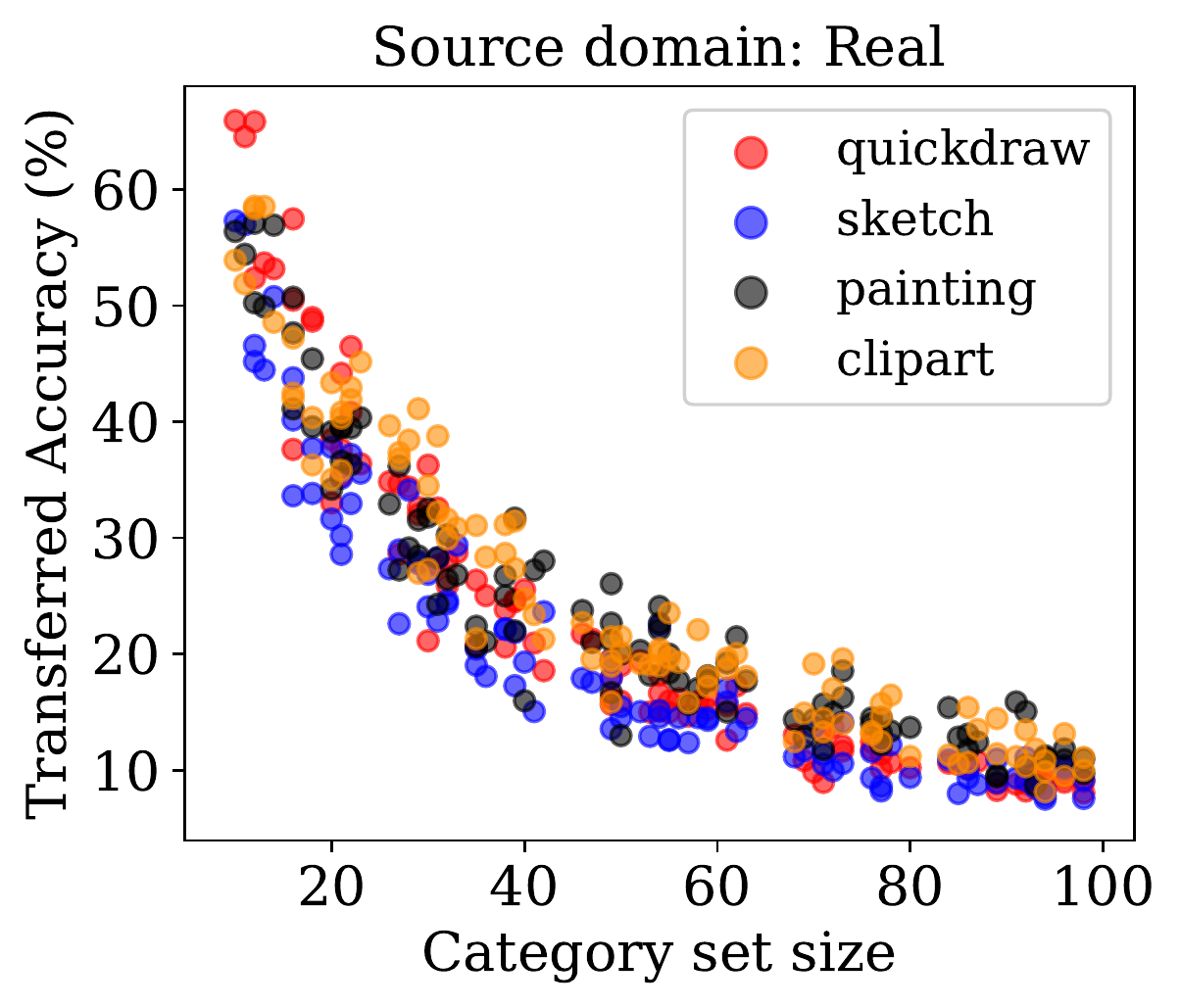}
  	\caption{Visualization of transfer accuracy v.s. category set size (number of categories). Points represent different target tasks.}
  	\label{fig:study of category set size}
  \end{figure}

Observations from our transferability experiments indicate that transfer accuracy drops when category set size (number of categories) increases, as shown in Figure \ref{fig:study of category set size}. A larger category set generally makes it more difficult to learn the target task well under the same training setting. 
  Such differences in the intrinsic complexity of the target task tends to overshadow the more subtle variations in transferability due to task and domain relatedness. Thus we make the quantitative comparisons under the fixed $category\_set\_size=50$ setting (Section 4.3 in paper) to show that our OTCE score is   capable of capturing these subtle variations.

%% file: appendix/tab_comparison.tex
\begin{table*}[!ht]
\begin{minipage}{\linewidth}

\footnotesize

\caption{Quantitative comparisons evaluated by Pearson correlation coefficients of transferability metrics and transfer accuracy through Retrain head under cross-domain cross-task transfer settings. Superscript $^*$ denotes $p>0.001$.}
\label{tab: OT-based NCE}

\centering

\begin{tabular}{ccccccccc}
\toprule

\multirow{3}{*}{Dataset}  & \multicolumn{3}{c}{Experimental setting}& \multirow{3}{*}{\best{OTCE}} & \multirow{3}{*}{OT-based NCE} &\multirow{3}{*}{LEEP\cite{LEEP}} & \multirow{3}{*}{NCE\cite{NCE}} & \multirow{3}{*}{H-score\cite{bao2019information}} \\
\cmidrule{2-4} 
 & Source domain & Target domain & Data property &  &  & & & \\
\midrule

\multirow{5} {*}{DomainNet} & C & P,Q,R,S & standard & \best{0.969} &\textbf{0.960} & 0.919 & 0.787 & 0.864  \\
& P & C,Q,R,S&  standard & \best{0.968} &\textbf{0.952}  & 0.886 & 0.812 & 0.858 \\
& Q & C,P,R,S& standard & \best{0.963} & \textbf{0.963} & 0.942 & 0.935 & 0.843 \\
& R& C,P,Q,S& standard & \best{0.972} & \textbf{0.951} & 0.892 & 0.851 & 0.870  \\
& S & C,P,Q,R& standard & \best{0.960} &\textbf{0.959} & 0.952 & 0.954& 0.882  \\
\cmidrule{1-9} 
\multirow{3} {*}{Office31} & A & D,W& standard &\best{0.829} &\textbf{0.813} & 0.805 & 0.796 & 0.590 \\
& D & A,W& standard  & \best{0.880} & 0.843 & \textbf{0.857} & 0.849 & 0.441 \\
& W & A,D& standard & \best{0.863} & 0.803 & \textbf{0.811} & 0.804 & 0.489 \\
\cmidrule{4-9} 
& & & \textbf{average}  & \best{0.926} &\textbf{0.906} & 0.883 & 0.849 & 0.730  \\

\midrule
\midrule

\multirow{6} {*}{DomainNet} & C & P,Q,R,S & fixed category set size & \best{0.757} & \textbf{0.729} & 0.614 & 0.535 & -0.599  \\
& P & C,Q,R,S&  fixed category set size & \best{0.712} & \textbf{0.647} & 0.480 & 0.418 & -0.541 \\
& Q & C,P,R,S& fixed category set size & \best{0.352} & \textbf{0.306} & 0.213 & 0.269 & -0.288 \\
& R& C,P,Q,S&fixed category set size & \best{0.639} & \textbf{0.587} & 0.465 & 0.440 & $-0.100^*$  \\
& S & C,P,Q,R&fixed category set size  & \best{0.435} &\textbf{0.443} & 0.381 & 0.427& -0.302  \\

\cmidrule{4-9} 
& & & \textbf{average}  &\best{0.579} & \textbf{0.542} & 0.431 & 0.418  &  -0.346 \\

\midrule
\midrule

\multirow{5} {*}{DomainNet} & C & P,Q,R,S & few-shot & \best{0.920} &\textbf{0.907} & 0.843 & 0.713 & 0.767  \\
& P & C,Q,R,S&  few-shot & \best{0.924}& \textbf{0.895} & 0.812 & 0.737 & 0.807 \\
& Q & C,P,R,S& few-shot & \best{0.852} & \textbf{0.857} & 0.836 & 0.825 & 0.786 \\
& R& C,P,Q,S& few-shot & \best{0.937} & \textbf{0.877} & 0.787  & 0.744& 0.814  \\
& S & C,P,Q,R& few-shot  & \best{0.922} & \textbf{0.901} & 0.886 & 0.884 & 0.834  \\
\cmidrule{1-9} 
\multirow{3} {*}{Office31} & A & D,W& few-shot  & \best{0.840} & \textbf{0.826} & 0.803 & 0.793 & 0.640 \\
& D & A,W& few-shot   & \best{0.933} & \textbf{0.931} & 0.923 & 0.930 & 0.413 \\
& W & A,D& few-shot  & \best{0.927} & \textbf{0.932} & 0.920 & 0.926 & $0.277^*$ \\
\cmidrule{4-9} 
& & & \textbf{average}  & \best{0.907} & \textbf{0.891} & 0.851 & 0.819  &  0.633 \\

\bottomrule

\end{tabular}

\end{minipage}
\end{table*}